# Steering Multistakeholder Negotiations Towards a Balanced Solution with LLM Agents: A Food Safety Case Study

Iurii Savvateev[1[0009-0000-5159-7239]], Josef Teichmann[2[0000-0003-0125-7484]], Matthias Filter[3[0000-0001-9347-021X]], Sophia Johler[1[0000-0003-4299-5651]], Alexandra Fetsch[1[0009-0006-8060-0314]]

[1] LMU Munich, Munich, Germany; Email: iurii.savvateev@lmu.de, sophia.johler@lmu.de, alexandra.fetsch@lmu.de

[2] ETH Zurich, Zurich, Switzerland; Email: jteichma@math.ethz.ch

[3] German Federal Institute for Risk Assessment, Berlin, Germany; Email:, matthias.filter@bfr.bund.de

**Abstract.**

Risk negotiations require stakeholders to translate heterogeneous, plain-text arguments into agreements that balance conflicting interests, yet these agreements are not necessarily checked against more formal benchmarks. We present a large language model (LLM)-based pipeline that links natural-language negotiation to benchmarks from game theory. The pipeline (i) derives stakeholder preference profiles from public statements made during public preliminary discussions using structured output extraction with LLMs and constrained Monte Carlo sampling, (ii) identifies the minimal profile adjustment needed to make a Nash Bargaining Solution (NBS) feasible, and (iii) supports negotiations between LLM agents representing real stakeholders using shared Bayesian opponent models. These models translate public statements into quantitative evidence, estimate each negotiation package's acceptability to every party and suggest interactions that maximize information gain. We tested the pipeline in negotiations on a food safety criterion for *Campylobacter spp.* in broiler meat. In a previous in-person seminar, stakeholders reached consensus on this criterion, although their stated preferences did not admit NBS. The automated pipeline independently captured most participant-provided preferences (75 of 84) and reproduced this infeasibility. Subsequent adaptation analysis identified a narrow deadlock that could be resolved between two stakeholders and in the follow-up simulated negotiations, the agents reached the NBS in 13 of 28 runs. Analysis of the unsuccessful runs indicated that moderator influence and coalition formation, rather than insufficient information, prevented the agents from reaching NBS. The pipeline thus offers a traceable way to anticipate disagreement, quantify the compromises needed and guide negotiations towards a balanced solution.

**Keywords:** Large language models · LLM-based agents · Multiagent systems · Bayesian modelling · Food safety · *Campylobacter spp.* · Nash bargaining solution

## 1 Introduction

Risk negotiations increasingly require stakeholders to interpret large volumes of heterogeneous information, decide under practical time constraints and balance partially conflicting interests. This is particularly relevant for interdisciplinary problems, which involve trade-offs across multiple objectives and stakeholder groups

Computational support can help organize evidence, structure stakeholder preferences, explore pathways toward compromise and assess how a proposed agreement balances the interests of the parties [1].

Food-safety risk management exemplifies this challenge, because decisions must connect scientific evidence with public-health protection, feasibility of interventions, economic consequences and stakeholder acceptability [2]. *Campylobacter jejuni* and *C. coli* are the leading bacterial causes of foodborne gastroenteritis in Europe. In 2024, campylobacteriosis was the most frequently reported zoonosis in the European Union (EU) [3], with sequelae such as Guillain–Barré syndrome and reactive arthritis [4]. Broiler meat is the most important vehicle: handling, preparation and consumption of broiler meat account for an estimated 20–30% of cases, and the chicken reservoir for up to 50–80% [5]. Because the infectious dose is low, risk depends on pathogen concentration rather than mere presence on the product; still, zero risk is neither achievable nor a meaningful management target, and every control strategy implicitly accepts some residual risk [6,7]. The EU process hygiene criterion (PHC) follows this risk-focused logic: exceedances trigger corrective action in the process, not product withdrawal [8]. As case numbers have not declined since its introduction in 2018, a binding food safety criterion (FSC) is currently under debate. An FSC would require non-compliant batches to be withdrawn, recalled or treated, with side effects such as higher testing costs, disproportionate pressure on small and medium-sized enterprises and food waste that conflicts with the EU's 2030 commitments [9,10]. Because these benefits and costs are incommensurable and valued differently by authorities, industry, consumers and environmental organizations, and classical quantitative microbial risk assessment rarely incorporates such trade-offs [6], negotiation-based approaches in which stakeholders agree on a tolerable level of residual risk have been proposed [1].

To test such an approach, we organized a seminar [11] in which participants from academia, governmental institutions and industry used an AI-augmented negotiation pipeline to discuss an FSC for *Campylobacter* spp. in broiler meat under EU Regulation (EC) No 2073/2005[12]. LLM agents assigned to each stakeholder group engaged in multi-step negotiations and identified issues and candidate options; seminar participants curated and scored them based on their preferences, and a further round of agent negotiations searched for compromises. Participants of the seminar reached consensus and found the LLM assistance useful [11]. However, a consensus is not necessarily a mathematically balanced outcome. A subsequent search for a disagreement-adjusted Nash bargaining solution (NBS), which maximizes the product of the stakeholders' gains above their disagreement points [13], showed that no negotiation package exceeded every stakeholder's threshold: the expressed preferences, thresholds and options did not jointly permit an agreement with positive gains for all stakeholders.

Here, we extend this framework [11] into a fully automated pipeline that explicitly searches for a path toward NBS. By combining automatic scoring, adaptation analysis and Bayesian opponent modelling, the pipeline translates heterogeneous, publicly stated stakeholder positions into a quantitatively balanced agreement through a traceable process.

## 2 Methods

The pipeline comprises three steps: automatic scoring, adaptation analysis and agent negotiation with Bayesian opponent models. Below we outline the key steps whereas the code, prompts, agent traces, simulation details along with current limitations are available at https://github.com/IuriiSavvateev/Nash_equilbirum_with_LLMs

### 2.1 Case Structure and Nash Bargaining Solution

The negotiation relied on the decomposition of the general topic to a list of specific challenges (i.e. issues) and possible solutions (e.g. options) achieved in the seminar [11]: type and stringency of the microbiological criterion (issue A, four options), primary control focus along the supply chain (issue B, five options), cost-burden sharing and food-waste mitigation (issue C, five options), implementation scope (issue D, four options) and environmental footprint of the criterion (issue E, three options), yielding a set $X$ of 1,200 negotiation packages $x = (x_A, \ldots, x_E)$. Each stakeholder $i$ (consumers, food industry, food safety authorities, environmental NGOs) assigned integer issue weights $w_{ij}$ for each issue $j \in \{A, \ldots, E\}$ with $\sum_j w_{ij} = 100$ and option scores, signaling a solution preference, $0 \leq s_{ij}(o) \leq w_{ij}$, so that the utility of a negotiation package is

$$u_i(x) = \sum_j s_{ij}\left(x_j\right). \qquad (1)$$

With a disagreement point $d = 65$ for every party, the NBS [13] is defined as

$$x^* \in \arg \max_{x \in X:\ u_i(x) \geq d\ \forall i} \prod_i \left(u_i(x) - d\right). \qquad (2)$$

We verified that at least one package gives every party more than d, and where the original scores did not satisfy this, they were minimally adapted (see section 2.3).

### 2.2 Automatic Scoring

For each stakeholder, an independent LLM (GPT-5.5) used a custom prompt with structured output to convert the stakeholder's own public preliminary-discussion statements into a set $\mathcal{C}_i$ of ordinal constraints: relevant issues, issue-importance orderings and within-issue option preferences. Score tables $S = (w, s)$ were then sampled from a maximum-entropy distribution conditioned on these constraints,

$$S^{(k)} \sim p(S) \propto p_0(S)\, \mathbb{1}[S \vDash \mathcal{C}_i], \qquad k = 1, \ldots, 20\,000, \qquad (3)$$

where under $p_0$ the active issues are uniform over admissible subsets (relevant issues always active), issue weights are Dirichlet(1) on the 100-point budget, and option scores are uniform fractions of their issue weight. Each draw was rounded to an integer table, and draws whose rounding broke a constraint were discarded. The central 90% credible interval $[q_{0.05}, q_{0.95}]$ of each option score was compared with the participant-provided score, and the sample closest to the componentwise medians, $\hat{S}_i \in \operatorname{argmin}_k \| S^{(k)} - \operatorname{med}(S) \|_1$, served as the representative profile

### 2.3 Adaptation Analysis

We searched for integer weight changes $\Delta w_{ij}$, with $\sum_j \Delta w_{ij} = 0$ and zero-weight issues held fixed, that admit a negotiation package with $u_i(x) \geq 66$ for all $i$. Option scores were rescaled as $s'_{ij}(o) \approx s_{ij}(o)\left(w_{ij} + \Delta w_{ij}\right)/w_{ij}$, rounded to integers that preserve all rankings, ties and zeros. Among feasible adaptations, we selected the one minimising $\max_{ij}|\Delta w_{ij}|$ and then $\sum_{ij}\Delta w_{ij}^2$. The adaptation distance was the number of transferred points,

$$D = 1/2\sum_i\sum_j|\Delta w_{ij}|. \quad (4)$$

### 2.4 Agent Negotiation

Four LLM agents (GPT-5.5, high reasoning effort), one per stakeholder, negotiated over the adapted automatic profile using OpenAI Agents SDK. Each agent knew the shared case description, its own public position, its private score table and its threshold $d$, and saw the last six public statements. In each run, one stakeholder acted as moderator and opened with a negotiation package. After six and twelve public statements, the moderator proposed an interim and a final negotiation package. Votes were computed as follows: a party accepted a negotiation package if its true score was at least $d$, and a negotiation package was adopted if at least three parties accepted it. The moderator role rotated across runs (seven runs per stakeholder; 28 in total). No agent accessed another party's private scores.

### 2.5 Bayesian Opponent Models

Before each move, an agent could query a Bayesian model[14] of each other party, built only from public information and therefore shared by all agents. The prior comprised $K = 12\,000$ samples from $p(S)$ in Eq. (3), without integer rounding. A semantic extractor (GPT-5.5 with structured outputs) converted each public statement into evidence $e$: option preferences, $s(o) > s(o')$, and unconditional acceptances, $u(x) \geq d$, or rejections, $u(x) < d$. An item was kept only if its verbatim quote occurred in the statement; conditional statements and proposals were ignored. With a deterministic likelihood, the posterior after the evidence $E_t$ is

$$p(S \mid E_t) \propto p(S)\prod_{e\in E_t}\mathbb{1}[S \models e], \qquad K_t = \{k: S^{(k)} \models E_t\}, \qquad (5)$$

with a fallback on the prior if $K_t = \emptyset$. A query returned posterior mean scores with 90% credible intervals, the acceptance probability

$$\hat{P}(u(x) \geq d) = \frac{1}{|K_t|}\sum_{k\in K_t}\mathbb{1}\left[u^{(k)}(x) \geq d\right], \qquad (6)$$

and one recommended question with the highest expected information gain, i.e., the largest entropy $H(\pi) = -\pi\log_2\pi - (1-\pi)\log_2(1-\pi)$. The candidate questions were acceptance of the proposed negotiation package, with $\pi = \hat{P}(u(x) \geq d)$, and every within-issue option pair not yet compared, with $\pi = \hat{P}\big(s(o) > s(o')\big)$.

### 2.6 Evaluation

Final-vote negotiation packages were compared with $x^*$ (see Eq. (2)) by the Euclidean and Manhattan distances between utility vectors and by the Hamming distance $\sum_j \mathbb{1}\left[x_j \neq x_j^*\right]$. Learning was quantified by (i) the share of within-issue option pairs $\mathcal{P}$ with different true scores that the posterior means $\bar{s}$ order correctly,

$$A = \frac{1}{|\mathcal{P}|} \sum_{(o,o\prime)\in\mathcal{P}} \mathbb{1}\left[\operatorname{sgn}\big(\bar{s}(o) - \bar{s}(o')\big) = \operatorname{sgn}\big(s(o) - s(o')\big)\right], \qquad (7)$$

averaged over the four models, and (ii) the NBS rank, i.e., the share of negotiation packages with a lower joint acceptance probability than the NBS,

$$R = \frac{1}{|X|} \sum_{x\in X} \mathbb{1}\left[\pi(x) < \pi(x^*)\right], \qquad \pi(x) = \prod_{i} \hat{P}_i\left(u_i(x) \geq d\right), \qquad (8)$$

excluding statements about the NBS itself. Start and final-vote values were compared with paired t-tests within outcome groups and are reported as mean ± SEM over runs.

# 3 Results and Discussion

## 3.1 Automatic Scoring Reproduces Participant Preferences and a Narrow Deadlock

Most scores provided by participants (75 of 84) fell within the central 90% credible interval of the automatically sampled score distributions (Fig. 1a), supporting the feasibility of extracting scores from the preliminary discussion alone. However, neither the participant-provided nor the automatic representative profile admitted a negotiation package acceptable to all parties, so no NBS existed. This shows that, in real negotiations, initial preferences can rule out NBS from the start, and the fact that the automatic procedure reproduced this outcome further supports the qualitative agreement between participant-based and automatic scoring.

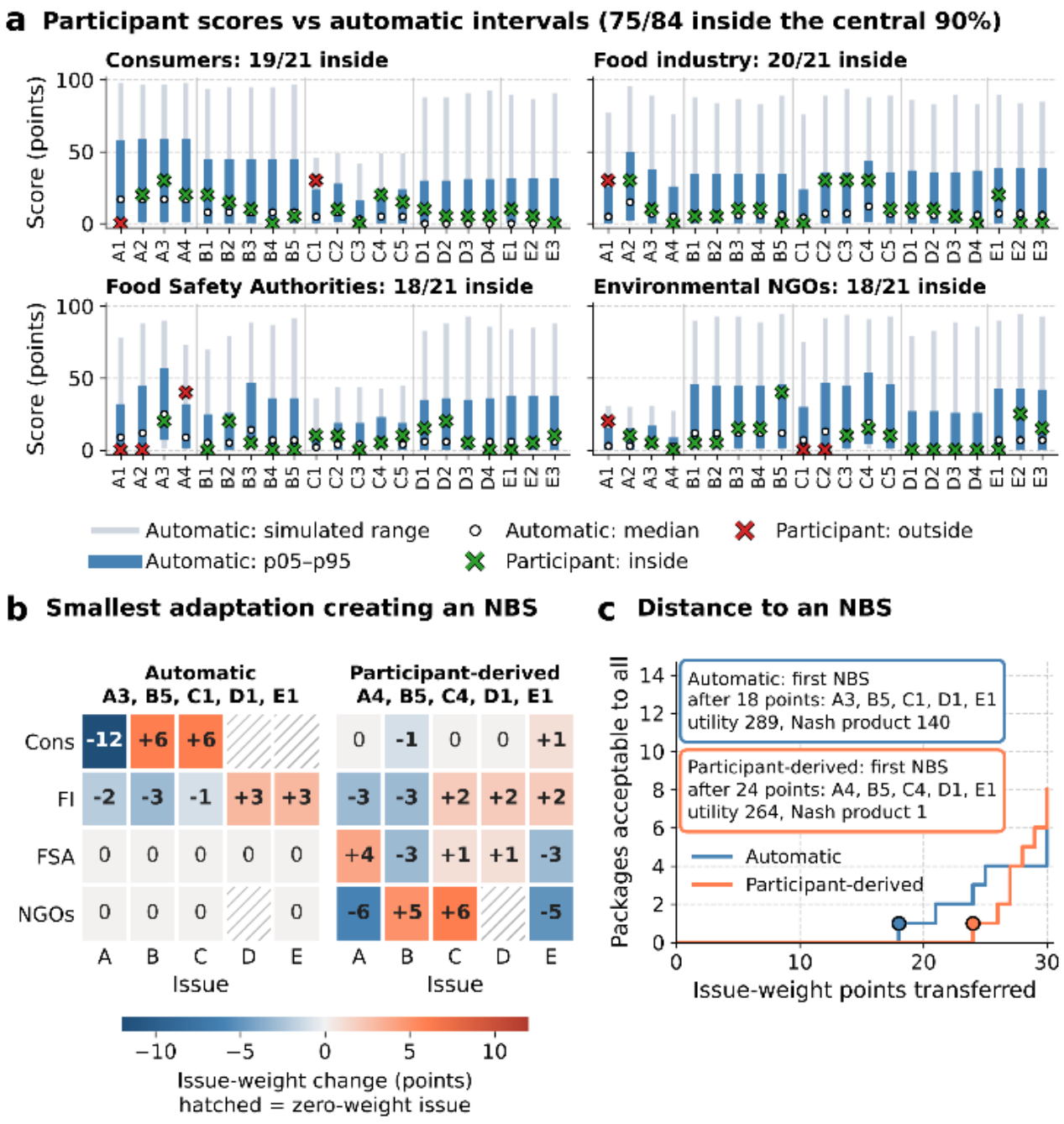

**Fig. 1.** Automatic versus participant-based scoring in the *Campylobacter* spp. case. **(a)** Participant scores (crosses) compared with the automatically generated score distributions per stakeholder and option: grey bars, full simulated range; blue bars, central 90% credible interval (p05–p95); white circles, medians. Green and red crosses lie inside and outside the interval, respectively. **(b)** Smallest integer adaptation of issue weights that gives every party at least 66 points (disagreement point 65) for the automatic (left) and participant-based (right) profiles. Cells give the change in issue weight (points of the 100-point budget); hatched cells are zero-weight issues, held fixed. **(c)** Number of negotiation packages acceptable to all parties versus the total issue-weight points transferred; circles mark the first NBS, boxes give its negotiation package, total utility and Nash product. Cons, consumers; FI, food industry; FSA, food safety authorities; NGOs, environmental NGOs.

The automatic landscape required fewer transferred points than the participant landscape to make an NBS feasible (18 vs. 24) and yielded an NBS with a higher total utility (289 vs. 264) and Nash product (140 vs. 1; Fig. 1b, c). For the *Campylobacter spp.* case, the deadlock was therefore narrow, and only two stakeholder groups (consumers, and food industry) had to adjust. Consumers scores were moved 12 points away from the type and stringency of the criterion (issue A) to the control focus (issue B) and cost-burden sharing (issue C), suggesting a moderate, risk-based criterion given stronger guarantees on cost allocation and waste reduction as a possible adjustment during the actual negotiations. The food industry profile was shifted by 6 points from issues A–C to implementation scope (issue D) and environmental footprint (issue E), suggesting a potential compromise on stringency, control focus and costs in exchange for the preferred terms on harmonization and environmental obligations as the most efficient suggestion to adapt the food industry profile. Most participant scores outside the credible interval also concerned issue A, marking the stringency of the criterion as the first point to clarify.

Automatic scoring can thus help stakeholders quantify their own preferences and anticipate the opponents' scores and disagreement points before formal negotiations.

### 3.2 LLM Negotiations with Bayesian Opponent Models Reach the NBS

With the adapted automatic profile (Fig. 1b), the agents reached the NBS (A3, B5, C1, D1, E1) in 13 of 28 runs and under every moderator (Fig. 2a), including runs whose opening negotiation package lay far away from it. The remaining runs ended either in agreements that passed with three votes while leaving one party below the acceptance threshold (8 runs), or without agreement (7 runs).

The Bayesian layer turned plain-text statements into quantitative evidence and actively shaped the discussion. In 48±4 % (error as SEM) public statements were recognized as explicit preferences or stances, and nearly all of these were consistent with the speaker's true scores (Fig. 2c, upper panel). The agents also used the model to guide the conversation through targeted public questions chosen to maximize the information gain (Fig. 2c, lower panel). Notably, the Bayesian models successfully learned in every run (Fig. 2d) leading to a statistically significant increase of the gained information in all outcome groups (Fig. 2e). The models also ranked the NBS by the final vote above most possible negotiation packages regardless of whether the NBS was eventually chosen or not (Fig. 2f), signalling that information about NBS was available in every run.

However, whether such NBS-related information led to the NBS depended strongly on the negotiation procedure (Fig. 2b). The moderator's agenda kept final proposals close to its opening negotiation package, while the three-of-four voting rule allowed majority coalitions to agree on negotiation packages that excluded one party, ending negotiations before the global optimum was reached. For the *Campylobacter spp.* case, all 21 agreements included a risk-based FSC with a moderate limit (A3) and an environment-first control focus (B5). Coalitions that missed the NBS left consumers behind on cost allocation (issue C) or the food industry behind on harmonization and environmental obligations (issues D and E), mirroring the adaptation in Fig. 1b. Failed negotiations mostly occurred under the food industry as moderator (5 of 7 runs), whose final proposals typically contained a lenient, phased FSC (A2) or departed from the environment-first focus and were blocked by the food safety authorities or consumers. In a real negotiation, the stringency of the criterion could therefore be settled early, with attention then focused on the accompanying measures, where a majority can most easily form an agreement at one party's expense.

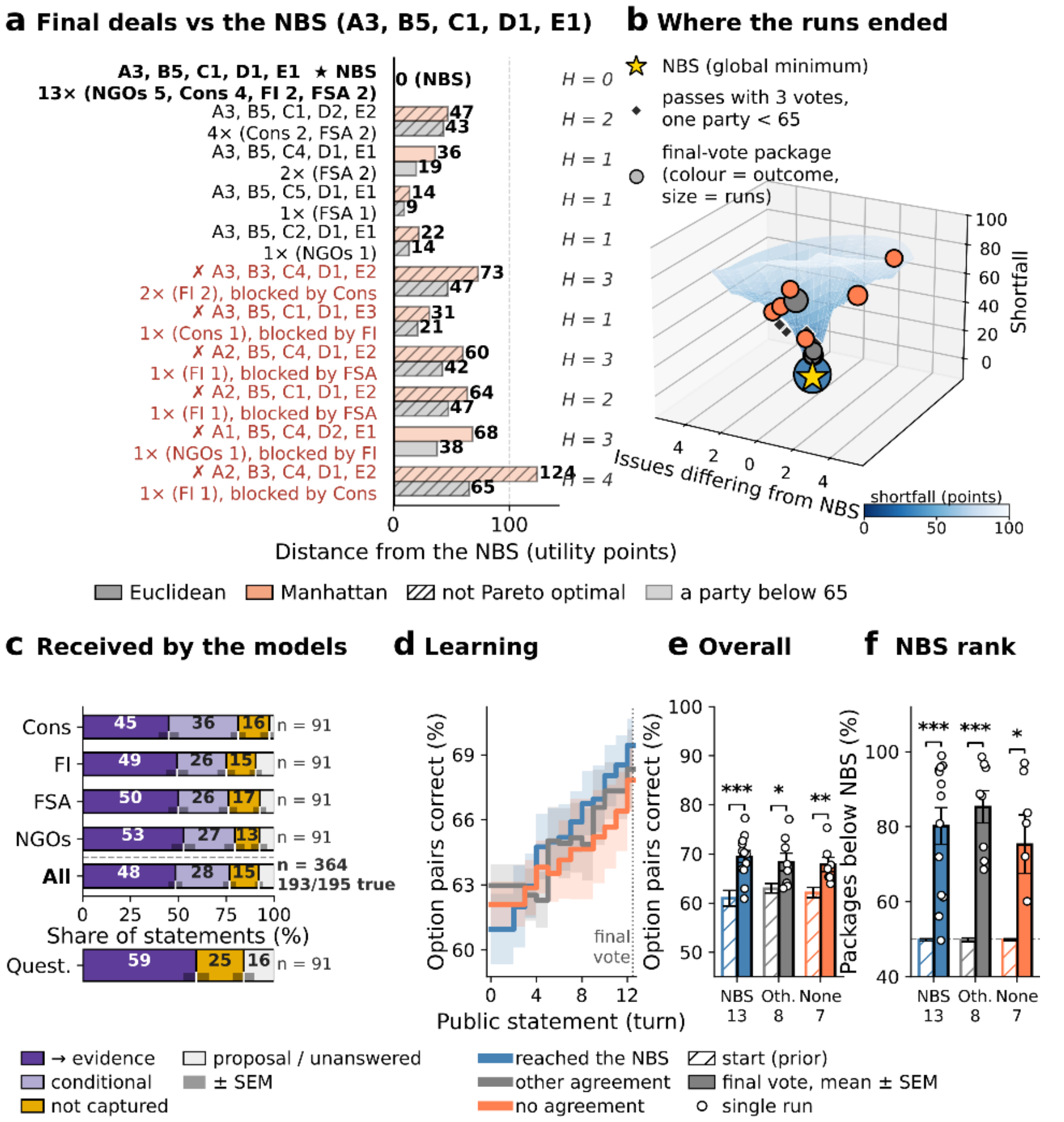

**Fig. 2.** Negotiation outcomes and perception of public statements by the Bayesian opponent models **(a)** Final-vote negotiation packages versus the NBS; rows give the negotiation package, number of runs and moderators; red rows (✗) were rejected, with the blocking party. Bars give the Euclidean (grey) and Manhattan (orange) distances of the negotiation package's utility vector from that of the NBS; H, Hamming distance. Hatched, not Pareto optimal; faded, at least one party below 65. **(b)** Negotiation landscape: negotiation packages arranged around the NBS (centre) by the number of differing issues (radius) and profile similarity (angle); height and colour give the total shortfall below 65 points summed over parties. Star, NBS (global minimum); diamonds, negotiation packages that pass with three votes while one party stays below 65; dots, final-vote negotiation package of each run (colour, outcome; size, number of runs). **(c)** Classification of public statements (upper bars; numbers in %) and questions (bottom bar) by the evidence extractor (mean share over runs; dark strips, ± SEM); n, number of statements; true, evidence consistent with the speaker's scores. **(d)** Share of within-issue option pairs ordered correctly by the posterior mean scores, averaged over the four opponent models (mean ± SEM over runs). **(e)** The same measure at the start and at the final vote. **(f)** Share of negotiation packages ranked below the NBS by joint acceptance probability at the start and at the final vote. In (e) and (f), numbers under the bars give the runs per outcome; *, **, ***: $p < 0.05, 0.01, 0.001$ (paired t-test, start vs. final vote). Cons, consumers; FI, food industry; FSA, food safety authorities; NGOs, environmental NGOs.

In summary, these results suggest that the automated pipeline can successfully identify a path towards NBS, as well as to highlight solutions that may satisfy the voting majority leaving minority parties behind. The latter is particularly useful to explore possible coalition formations when preparing for real-world negotiations.

# 4 Conclusion and outlook

## 4.1 An Autonomous and Traceable Path to the Nash Bargaining Solution

We developed a pipeline that automates the transition from plain-text public statements to a balanced agreement: it derives preference scores from a preliminary public discussion, quantifies the smallest preference adaptation that makes a NBS feasible and supports agent negotiations with Bayesian opponent models. In the *Campylobacter spp*. use case, participants reached consensus although their preferences admitted no NBS. The pipeline reproduced this infeasibility, identified a narrow deadlock resolvable by redistributing a few issue-weight points, and reached the resulting NBS under every moderator. Because every model update derives from an explicit public statement, each step can be traced and audited (see Github), which supports transparency and accountability in line with the aims of the EU AI Act[15] .

Importantly, negotiation runs that missed the NBS did not lack information, as the opponent models learned comparable information in all runs. Instead, deviations reflected the procedure: the moderator's agenda and majority coalitions under the three-of-four voting rule. Thus, procedural rules may matter as much as information exchange, and coalition analysis can help stakeholders anticipate agreements that would leave one party behind.

## 4.2 Generative AI Translates Plain-Text Negotiation into Game-Theoretic Terms

Game-theoretic concepts such as the NBS [13] require quantitative utilities, whereas real negotiations unfold in natural language. Generative AI can bridge this gap through semantic text analysis: an LLM extracted ordinal preferences that reproduced most participant-provided scores and converted about half of all public statements into evidence for the opponent models. These models, in turn, estimated each party's scores and the acceptability of every negotiation package, and suggested the most informative questions. Thus, LLM performs the semantic translation, while inference remains in a transparent, reproducible Bayesian model [14]. This combines the flexibility of language models with the auditability of formal models.

## 4.3 Limitations and Outlook

This study uses a single case with simulated agents, whose behaviour may differ from human negotiators. About a quarter of the statements were conditional and discarded. Extending the extractor to conditional trade-offs could accelerate learning. Whether real stakeholders would accept the adapted profiles must be tested with participants. Future work will integrate the pipeline into applicable real-world use cases[11] .

**Acknowledgments.** This work was conducted during the MicRISK2030 project funded by the SNSF (Swiss National Science Foundation, Sinergia grant number CVRSII5_209253). Additional funding was received by the German Federal Ministry of Food and Agriculture (BMEL) in the research project "KI- & Daten-Akzelerator (KIDA).

**Disclosure of AI usage and Interests.** The authors used AI-assisted tools to improve spelling, grammar, clarity and readability. AI-assisted tools were also used for the code preparation. After using the tools, the authors reviewed the materials and take full responsibility for the final content. Authors declare no competing interests.